\pdfoutput=1

\documentclass[11pt]{article}

\usepackage{acl}
\usepackage{times}
\usepackage{latexsym}

\usepackage[T1]{fontenc}

\usepackage[utf8]{inputenc}

\usepackage{microtype}

\usepackage{inconsolata}

\usepackage{graphicx}
\usepackage{breqn}
\usepackage{amssymb}

\usepackage{inconsolata}

\usepackage{bibentry}
\usepackage{dialogue}
\usepackage{times}
\usepackage{latexsym}
\usepackage{mathrsfs}
\usepackage{amssymb}
\usepackage{amsmath}
\usepackage{url}
\usepackage{pifont}
\usepackage{graphicx}
\usepackage{xcolor}
\usepackage{braket}
\usepackage{bbold}
\usepackage[utf8]{inputenc}
\usepackage[toc,page]{appendix}
\usepackage{cleveref}
\usepackage{booktabs,graphicx,makecell,siunitx,eqparbox}
\usepackage[utf8]{inputenc}
\usepackage{color,soul}
\usepackage{arydshln}
\usepackage{enumitem}
\usepackage{float}
\usepackage{colortbl}
\usepackage[most]{tcolorbox}
\definecolor{Red}{RGB}{224,102,102}
\definecolor{Green}{RGB}{147,196,125}
\definecolor{Blue}{RGB}{102,178,255}
				
\title{ALMol: Aligned Language-Molecule Translation LLMs through Offline Preference Contrastive Optimisation}

\author{Dimitris Gkoumas \\
  Queen Mary University of London, London, UK \\
  \texttt{d.gkoumas@qmul.ac.uk} \\}

\begin{document}
\maketitle
\begin{abstract}
The field of chemistry and Artificial Intelligence (AI) intersection is an area of active research that aims to accelerate scientific discovery. The integration of large language models (LLMs) with scientific modalities has shown significant promise in this endeavour. However, challenges persist in effectively addressing training efficacy and the out-of-distribution problem, particularly as existing approaches rely on larger models and datasets. In this context, we focus on machine language-molecule translation and deploy a novel training approach called contrastive preference optimisation, which avoids generating translations that are merely adequate but not perfect.  To ensure generalisability and mitigate memorisation effects, we conduct experiments using only 10\% of the data. Our results demonstrate that our models achieve up to a 32\% improvement compared to counterpart models.  Finally, we introduce a fine-grained, domain-agnostic evaluation method to assess hallucination in LLMs and promote responsible use.

\end{abstract}
\section{Introduction}
The world is facing unprecedented complexity in the form of global challenges such as climate change, healthcare, and pandemics. Innovative scientific solutions are urgently needed to address these challenges. Chemistry has been at the forefront of developing such solutions, pioneering new drugs~\cite{ferguson2018kinase}, creating advanced materials~\cite{kippelen2009organic}, or enhancing chemical processes~\cite{zhong2023reaction}. However, these frontiers are vast and require the involvement of Artificial Intelligence (AI) technology to navigate them effectively. 

Large language models (LLMs) have shown promising potential for accelerating scientific discovery across various domains, including chemistry, biology, and materials science~\cite{zhang2307artificial,ai4science2023impact}. Existing work has applied successful paradigms from natural language processing (NLP) and multimodal representation learning to the chemistry domain. One common approach involves converting the inherent three-dimensional structures of molecules into SMILES, which provide a mapping to symbolic character-level representations. Subsequently, researchers have explored learning language-molecule representations either in separate yet coordinated spaces~\cite{edwards2022translation,edwards2021text2mol,liu2023multi}, in a joint space~\cite{liu2023molxpt}, or through hybrid approaches~\cite{luo2023molfm,christofidellis2023unifying}. In light of the recent significant advancements in the field, none of the above approaches effectively tackle the inherent challenges in training such models. Instead, they rely on sparse or noisy synthetic data, often necessitating exponentially more data than is typically used in NLP tasks~\cite{edwards2024building}.

However, training on larger models and datasets does not necessarily guarantee higher performance. 
A successful paradigm that augments the capabilities of LLMs across multiple NLP tasks is Reinforcement Learning with Human Feedback (RLHF)~\cite{ouyang2022training}. Although initially challenged by issues of slowness and instability, recent research has addressed many of these challenges by shifting towards closed-form losses that operate directly on offline preference data~\cite{rafailov2024direct}. RLHF has demonstrated superior performance compared to standard minimising cross-entropy optimisation approaches.

In this context, we address challenges related to effectively training robust language models when integrated with scientific modalities.  We deploy a novel way of training LLMs for language-molecule translation that avoids generating translations that are only adequate but not perfect, called contrastive preference optimisation (CTO)~\cite{xu2024contrastive}. CTO is based on offline preferences instead of supervised fine-tuning, mimicking reference translations. To ensure that our models can effectively generalise instead of memorising patterns, we conduct experiments using only 10\% of the L+M-24 dataset~\cite{edwards2024building}. Our contributions have as follows:
\begin{itemize}[noitemsep,topsep=0pt,parsep=0pt,partopsep=0pt,leftmargin=*]
    \item Our models achieve significant performance improvements across various evaluation metrics compared to models trained on extensive in-distribution and out-of-distribution data ($\S~\ref{sec:exp_results}$).
    \item We showcase their robustness through experiments comparing pivot and minor cross-modals. Our empirical results demonstrate that our models consistently outperform the leading baseline, Meditron, which is trained on the entire dataset, even in agnostic cross-modal scenarios ($\S~\ref{sec:exp_results}$).
    \item We propose a fine-grained evaluation method that is domain-independent, assessing factual consistency in generated captions using a question-answering evaluation metric and measuring overlaps of unigrams in generated molecules against references ($\S~\ref{sec:eval_method}$). Our analysis shows that our models achieve improved factual consistency and character-level n-gram overlaps for caption and molecule generation ($\S~\ref{sec:eval_results}$).
\end{itemize}

\section{Background}
\label{sec:background}

Reinforcement Learning with Human Feedback (RLHF) optimisation~\cite{ouyang2022training}  operates with a triple dataset $\mathcal{D}= \{x, y_w, y_l\}$, where $y_w$ and $y_l$ represent preferred and dis-preferred outputs, corresponding to input $x$, such that $y_w \succ y_l$ for $x$. The probability of  $y_w$ over $y_l$ in pairwise comparisons is typically computed using the Bradley-Terry model~\cite{bradley1952rank}:
\begin{equation}
    p^* (y_w \succ y_l | x) = \sigma (r^*(x,y_w)-r^*(x,y_l))
\end{equation}
where $\sigma$ is the logistic function, and $r^*$ denotes the reward function that underlies the preferences.

As obtaining the reward directly from a human would be prohibitively expensive, a reward model $r_\phi$ is trained to act as a surrogate by minimising the negative log-likelihood of the preference data:
\begin{equation}
    \mathcal{L} (r_\phi) =  -\mathbb{E}_{(x,y_w,y_l)\sim \mathcal{D}} [\log \sigma (r_\phi(x,y_w)-r_\phi(x,y_l))] 
\end{equation}
Additionally, the Kullback-Leibler (KL) divergence between the outputs generated by $\pi_{\text{ref}}$ and the parameterised $\pi_{\theta}$ models serves as an additional reward signal, ensuring that the generated responses closely align with the reference model. Consequently, an optimal model $\pi_{\theta}$ is one that maximises:
\begin{equation}
    \mathbb{E}_{(x \in \mathcal{D}, y \in \pi_{\theta})}[r_{\phi}(x,y)] - \beta \mathcal{D}_{\text{KL}}\bigr(\pi_{\theta}(y|x) || \pi_{\text{ref}}(y|x)\bigr)
\label{eq:rlhf}    
\end{equation}
where $\beta$ is the temperature parameter typically $\in [0.1, 0.5]$. 

RLHF can present challenges due to its inherent slowness and instability, especially in distributed settings~\cite{zheng2024balancing}. Recent work has shifted towards closed-form losses to align LLMs with human preferences.  Here, we experiment with contrastive preference optimisation that adopts a closed-form loss for RLHF.

\section{Methodology}

\subsection{Task Formulation}


Let $(x,y)$ be a pair of source and target sequences mapped to $\mathrm{X}$ and $\mathrm{Y}$ spaces, respectively. We cast the problem of language-molecule translation as a cross-modal translation task that operates on offline preference data $\mathcal{D}=\{x^{(i)},y_w^{(i)},y_l^{(i)}\}_{i=1}^N$, where $x$  is an input,  $y_w$ are preferred (e.g. human gold standard) and $y_l$ dis-preferred outputs (typically synthetic, obtained from an appropriate translation model), and  $N$ is the total number of pairs. The goal is to learn an optimal function  $f: X \leftrightarrow Y$ through a model $\pi_{\theta}$ parameterised by \(\theta\). We coordinate the two spaces through instructional modelling to regulate the translation process in both directions. Specifically, for LMolT, we use instructions for language-to-molecule and molecule-to-language translation (see Appx.~\ref{appx:instructions}).

\subsection{Contrastive Preference Optimisation}

Contrastive preference optimisation (CTO)~\cite{xu2024contrastive} addresses challenges stemming from the inherent limitation in RLHF, as discussed in $\S$~\ref{sec:background}, and from the necessity of high-quality data. CTO is  a general approximation of Eq.~\ref{eq:rlhf}  using a uniform reference model, which assumes equal likelihood for all possible generated outputs:
\begin{align}
    &\mathcal{L}(\pi_{\theta};U) = -\mathbb{E}_{(x,y_w,y_l)\sim D} \nonumber \\
    &\Biggl[ \log \sigma \Bigl(\beta \log \pi_{\theta}(y_w|x) - \beta \log \pi_{\theta}(y_l|x)\Bigr) \Biggr]
    \label{eq:cpo}
\end{align}
where $\pi_{\theta}$ is parameterised model by $\theta$ and $\beta$ hyper-parameter (please refer $\S$~\ref{sec:background}). Eq.~\ref{eq:cpo} implies that the loss is calculated based on how well the generated translations match this uniform distribution of possible translations, rather than being biased towards any particular translation. To maintain $\pi_{\theta}$ close to the preferred data distribution, a behaviour cloning (BC)~\cite{hejna2023contrastive} regulariser is introduced:
\begin{align}
&\min_{\theta} \mathcal{L}(\pi_{\theta}, U) \quad \text{s.t.}  \quad \nonumber \\ &\mathbb{E}_{(x,y_w)\sim D} \Bigr[ \mathbb{KL}\bigl(\pi_w(y_w|x) || \pi_{\theta}(y_w|x)\bigl) \Bigr] < \epsilon,
\end{align}
Here, $\epsilon$ denotes a small positive constant, and $\mathbb{KL}$ signifies the Kullback-Leibler divergence. The regulariser is enhanced with an additional SFT term on the preferred data, bolstering the CPO loss as:

\begin{equation}
\mathcal{L}_{\text{CPO}} = \min_{\theta} \underbrace{\mathcal{L}(\pi_{\theta}, U)}_{\mathcal{L}_{\text{prefer}}} - \underbrace{\mathbb{E}_{(x,y_w)\sim D} \bigr[\log \pi_{\theta}(y_w|x)\bigr]}_{\mathcal{L}_{\text{NLL}}}
\end{equation}

\subsection{Proposed Evaluation Methodology}
\label{sec:eval_method}
Prior studies have utilised embedding representations, for assessing the semantics in chemical-domain models~\cite{jaeger2018mol2vec,edwards2021text2mol,christofidellis2023unifying}. However, these approaches require domain adaptation for out-of-distribution data~\cite{edwards2024building} and might lead to opaque and arbitrary outcomes~\cite{steck2024cosine}. We address these limitations by introducing a scalable fine-grained evaluation methodology for assessing the presence of hallucinations\footnote{Hallucination in LLMs refers to a phenomenon where the generated outputs are inaccurate, nonsensical, or contradictory to the provided factual information.} in generated outputs.

\begin{figure}[hbt!]
\centering
\includegraphics[width=.5\textwidth]{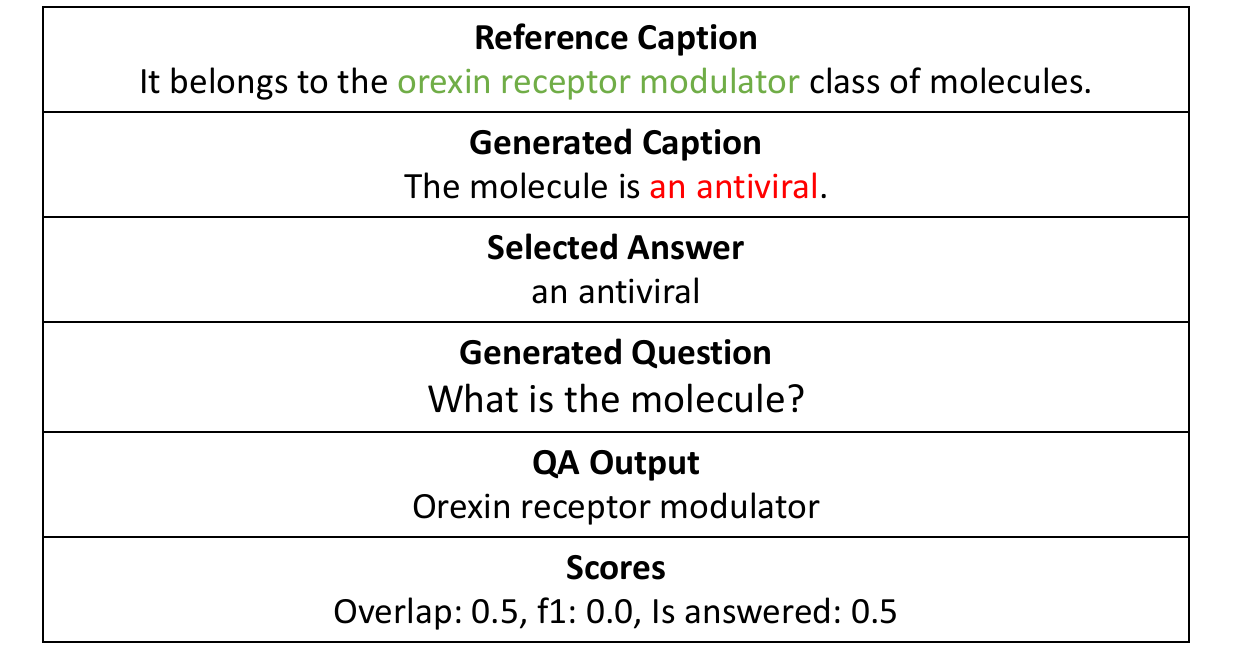}
\caption{A toy example illustrating a factual inconsistency between a generated and a reference caption. The QAFactEval metric selects a noun-phrase answer from the generated caption. A QG model then generates an associated question that a QA model answers based on the reference caption. The scores measure the semantic overlap between the QA model's answer and the selected answer from the generated caption} 
\label{fig:qafacteval}
\end{figure}

\paragraph{Language Evaluation:} For molecule-to-language translation, we deploy the QAFactEval~\cite{fabbri2022qafacteval} metric to evaluate the factual consistency of generated captions. QAFactEval first selects noun phrases and named entities (NER) from the generated outputs. A question generation (QG) model then formulates associated questions, which a question answering (QA) model addresses based on the reference text. QAFactEval measures the semantic overlap between the QA model's responses and the selected answers to produce the final metric score. An example is illustrated in Fig.~\ref{fig:qafacteval}. Here, we report the semantic \textit{overlap}, the \textit{$f_1$ accuracy}  between the QA model and the selected answer, and \textit{answerability}, which is the probability of the question being answered by the reference caption.

\paragraph{Molecule Evaluation:} For language-to-molecule translation, we employ the Chr-F metric, an F-score statistic, to evaluate character n-gram matches between prediction-reference pairs \cite{popovic2015chrf}. This metric assesses the matches in generated molecules against their references by averaging the scores of unigram, bigram, and trigram matches. A higher Chr-F score indicates better performance.

\paragraph{Bias Evaluation:} We also calculate the character and token length bias in generated-reference pairs of molecules and captions, respectively, to investigate potential length bias in the evaluated LLMs.

\section{Experiments}

\subsection{Data}
We conduct experiments on the \textit{L+M-24} benchmark dataset, which encompasses both molecule and linguistic modalities~\cite{edwards2024building}. It is divided into four categories, each with significant applications in small-molecule domain; biomedical; light and electricity; human interaction and organoleptics; and agriculture and industry. The training and validation subsets consist of  approximately 127k and 34k language-molecule pairs, respectively. Here, we utilise 10\% of these subsets for training and validation. To operationalise CTO, we recreate a triples dataset consisting of preferred and dis-preferred outputs (see $\S$~\ref{sec:background}), where the former are the golden references and the latter are generated from MolT5~\cite{edwards2022translation}. For evaluation, we randomly selected 3k unseen pairs from a distinct dataset provided by the research group of L+M-24.\footnote{Sampling is conducted from a distinct subset.}

\subsection{Bechmark Models}
We compare our results with established language-molecule models as captured in the literature:
\begin{itemize}[noitemsep,topsep=0pt,parsep=0pt,partopsep=0pt,leftmargin=*]
\item TxtChem-T5~\cite{christofidellis2023unifying}: A T5 model trained on both linguistic and molecule modalities with a multi-task objective across various datasets, including the CheBI-20 dataset~\cite{edwards2022translation}, akin to \textit{L+M-24}.
\item Chem-LLM~\cite{zhang2024chemllm}: An InternLM2-Base-7B model, trained on an extensive chemical domain knowledge dataset, with the direct preference optimisation objective~\cite{rafailov2024direct}, achieves results comparable to GPT-4.
\item Meditron~\cite{chen2023meditron}: A Meditron-7B model fine-tuned on the entire \textit{L+M-24} for unidirectional language-molecule translation.
\item SFT-Meditron: We fine-tune Meditron-7B on a 10\% subset of \textit{L+M-24} for bi-directional machine language-molecule translation.
\end{itemize}

\subsection{Experimental Settings}
Here, we train Meditron with CTO on a 10\% subset of \textit{L+M-24}. We experiment with both language and molecule weight initialisation obtained from Meditron trained on the entire data~\cite{edwards2024building}. We refer to them as CTO-Meditron$_{\overrightarrow{Lan.}}$ and CTO-Meditron$_{\overrightarrow{Mol.}}$, respectively. We train the models with QLoRA~\cite{dettmers2024qlora}. For evaluation, we adopt established metrics in~\cite{edwards2022translation}.

\subsection{Experiment Results}
\label{sec:exp_results}
Table~\ref{tbl:mol2cap} presents a summary of the molecule-to-language results. We observed a significant decrease in performance for benchmark models trained on extensive data with SFT when tested on out-of-distribution data. Among the baseline models, Meditron demonstrated the highest performance, likely due to its training on the entire \textit{L+M-24} dataset utilised in our experiments. Training Meditron with SFT for bi-directional language-molecule translation has demonstrated neither effectiveness (see Table~\ref{tbl:cap2mol}) nor efficiency (refer to Appx.~\ref{appx:effect_effic}). This suggests that the performance in our experiments is not dependent on memorised patterns from Meditron trained on the entire dataset. In contrast, our models trained with the CTO objective on only 10\% of \textit{L+M-24} achieved a remarkable improvement in performance across diverse evaluation metrics, up to 32\% compared to Meditron trained on the entire dataset. This improvement is consistent, as our model consistently enhances performance when initialised from agnostic cross-modals, i.e.,  CTO-Meditron$_{\overrightarrow{Lan}}$ in Table~\ref{tbl:cap2mol}.

\begin{table*}[ht]
  \centering
 
  \resizebox{\textwidth}{!}{%
  \begin{tabular}{ccccccc }
    \toprule

    \textbf{Model} & $\textbf{Blue-2} \uparrow$ & $\textbf{Blue-4} \uparrow$ & $\textbf{Rouge-1} \uparrow$  &   $\textbf{Rouge-2} \uparrow $ &  $\textbf{Rouge-L} \uparrow$  & $\textbf{METEOR} \uparrow$     \\ \hline

    TxtChem-T5 & 0.08  & 0.09  &  0.19  & 0.06 &  0.17 & 0.16\\ \hline
    Chem-LLM & 0.03 & 0.00  &  0.11  & 0.02 & 0.09 & 0.14 \\ \hline
    Meditron &  0.42 &  0.30 &  0.63  & 0.47 & 0.49 & 0.54\\  \hline
    SFT-Meditron & 0.37 & 0.26  &  0.54  & 0.39 & 0.38&  0.60\\ \hline
    CTO-Meditron$_{\overrightarrow{Lan}}$ & 0.62 (\textcolor{Green}{+0.20}) & 0.45  (\textcolor{Green}{+0.15})  & 0.67  (\textcolor{Green}{+0.03})  & 0.50  (\textcolor{Green}{+0.03})& 0.48  (\textcolor{Red}{-0.01}) & 0.62  (\textcolor{Green}{+0.08})\\ \hline
    CTO-Meditron$_{\overrightarrow{Mol}}$ & 0.74 (\textcolor{Green}{+0.32}) & 0.53 (\textcolor{Green}{+0.23})  & 0.76 (\textcolor{Green}{+0.10})  & 0.56 (\textcolor{Green}{+0.09}) & 0.53 (\textcolor{Green}{+0.04}) & 0.71(\textcolor{Green}{+0.17})\\ 
  \bottomrule
  \end{tabular}
 }
  \caption{Molecule-to-language translation results.  Arrows next to metrics indicate the higher value the better performance. Numbers in parentheses show deviations from Meditron trained on the entire dataset.}
  \label{tbl:cap2mol}
\end{table*}

We observed similar performance patterns for language-to-molecule translation as reported in Table~\ref{tbl:mol2cap}. However, even though our model achieved better performance compared to Meditron when initialised from agnostic cross-modals, it struggled to learn molecular patterns (see CTO-Meditron$_{\overrightarrow{Mol.}}$ in Table~\ref{tbl:mol2cap}). This suggests that language plays a pivotal role in the molecule modality. In the future, we aim to explore more advanced initialised methods to address this challenge.

\begin{table*}[ht]
  \centering
 
  \resizebox{\textwidth}{!}{%
  \begin{tabular}{ccccccccc }
    \toprule

    \textbf{Model} & $\textbf{BLEU} \uparrow$ & $\textbf{Exact} \uparrow$ & $\textbf{Levenshtein} \downarrow$  &   $\textbf{MACCS FTS} \uparrow $ &  $\textbf{RDK FTS} \uparrow$  & $\textbf{Morgan FTS} \uparrow$ & $\textbf{FCD} \downarrow$ &    $\textbf{Validity} \uparrow$  \\ \hline

   TxtChem-T5 & 0.18 &  0.00 &  133.29  & 0.21 & 0.10 & 0.03 &  37.67 & 0.58\\ \hline
    Chem-LLM  &  0.04 & 0.00  &  732.74 &  0.00 &  0.00  & 0.00   & 59.44  & 0.19 \\ \hline
    Meditron & 0.43 & 0.00  &  66.16  & 0.35 & 0.29  & 0.19 & 13.64 & 0.57\\ \hline
    SFT-Meditron & 0.30 & 0.00  & 186.99   & 0.70  & 0.62&  0.41& 11.14 &0.98 \\ \hline
    CTO-Meditron$_{\overrightarrow{Lan.}}$ & 0.71 (\textcolor{Green}{+0.28}) & 0.00  &  42.65 (\textcolor{Green}{-23.51})  & 0.78 (\textcolor{Green}{+0.43}) & 0.70 (\textcolor{Green}{+0.41})& 0.48 (\textcolor{Green}{+0.29}) & 4.19 (\textcolor{Green}{-9.45}) & 1.00 (\textcolor{Green}{+0.43}) \\ \hline
    CTO-Meditron$_{\overrightarrow{Mol.}}$ & 0.52 (\textcolor{Green}{+0.09})& 0.00  &  76.95 (\textcolor{Red}{+10.43})  & 0.52 (\textcolor{Green}{+0.17}) & 0.49 (\textcolor{Green}{+0.20})& 0.37 (\textcolor{Green}{+0.18}) & 27.39 (\textcolor{Red}{+13.75})& 0.58 (\textcolor{Green}{+0.01}) \\

  \bottomrule
  \end{tabular}
 }
  \caption{Language-to-molecule translation results. Arrows next to metrics  indicate whether higher or lower values denote better performance. Numbers in parentheses show deviations from Meditron trained on the entire dataset.}
  \label{tbl:mol2cap}
\end{table*}

\subsection{Evaluation Results}
\label{sec:eval_results}

\begin{figure}[hbt!]
\centering
\includegraphics[width=.5\textwidth]{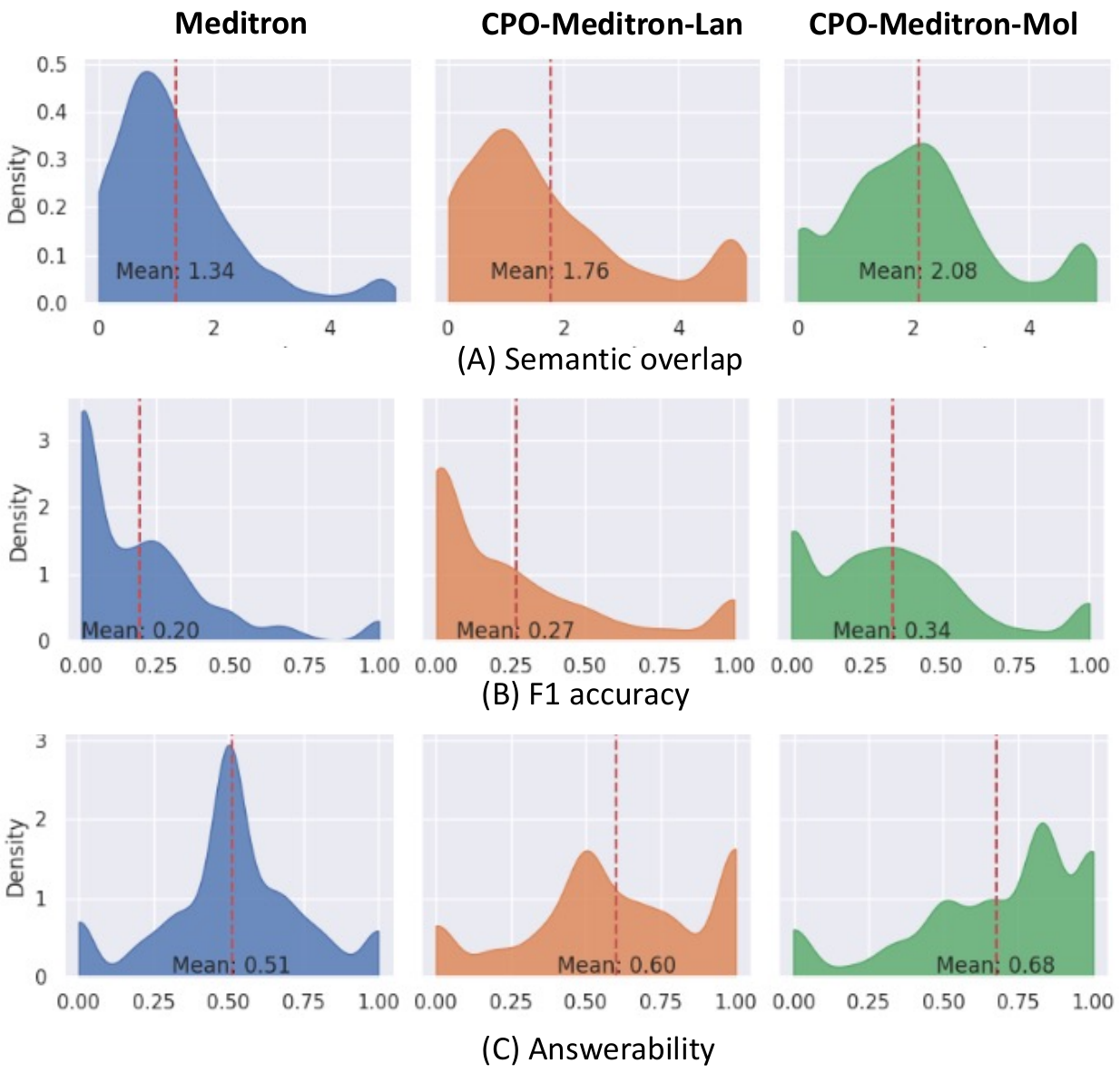}
\caption{Factual consistency in generated captions against references, assessed through (A) semantic overlap, (B) F1 accuracy, and (C) answerability using QAFactEval ($\S$~\ref{sec:eval_method}) across various LLMs.} 
\label{fig:language_factuality}
\end{figure}

Fig.~\ref{fig:language_factuality} illustrates the evaluation results on the factual consistency of generated captions against references for the molecule-to-language task.  CTO-Meditron$_{\overrightarrow{Mol.}}$, trained on 10\% of the available data, exhibited superior factual consistency, achieving a semantic overlap of 2.08, $f1$ accuracy of 0.34, and answerability of 0.68, compared to 1.34, 0.20, and 0.51, respectively, for Meditron trained on the entire dataset.  CTO-Meditron$_{\overrightarrow{Lan.}}$ also outperformed Meditron but showed lower performance than  CTO-Meditron$_{\overrightarrow{Mol.}}$. We attribute this to the model being initialised by agnostic cross-modals.

\begin{figure}[hbt!]
\centering
\includegraphics[width=.5\textwidth]{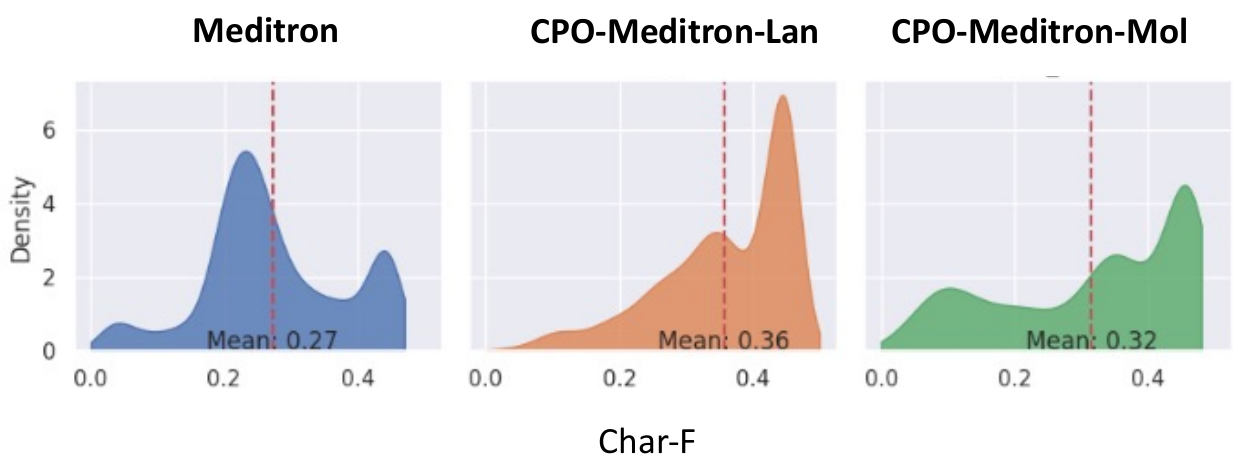}
\caption{Overlaps of n-gram matches between generated and reference molecules as captured by the char-F ($\S$~\ref{sec:eval_method}) score across various LLMs.} 
\label{fig:molecule_charf}
\end{figure}

For the language-to-molecule task, we observed that both Meditron$_{\overrightarrow{Lan.}}$ and Meditron$_{\overrightarrow{Mol.}}$ achieved similar performance in terms of uni-, bi-, and tri-gram overlaps between generated and reference pairs, outperforming Meditron (see Fig.~\ref{fig:molecule_charf}). However, when the model was initialized with known cross-modal weights, i.e., Meditron$_{\overrightarrow{Lan.}}$, it achieved a slightly increased performance

\begin{figure}[hbt!]
\centering
\includegraphics[width=.5\textwidth]{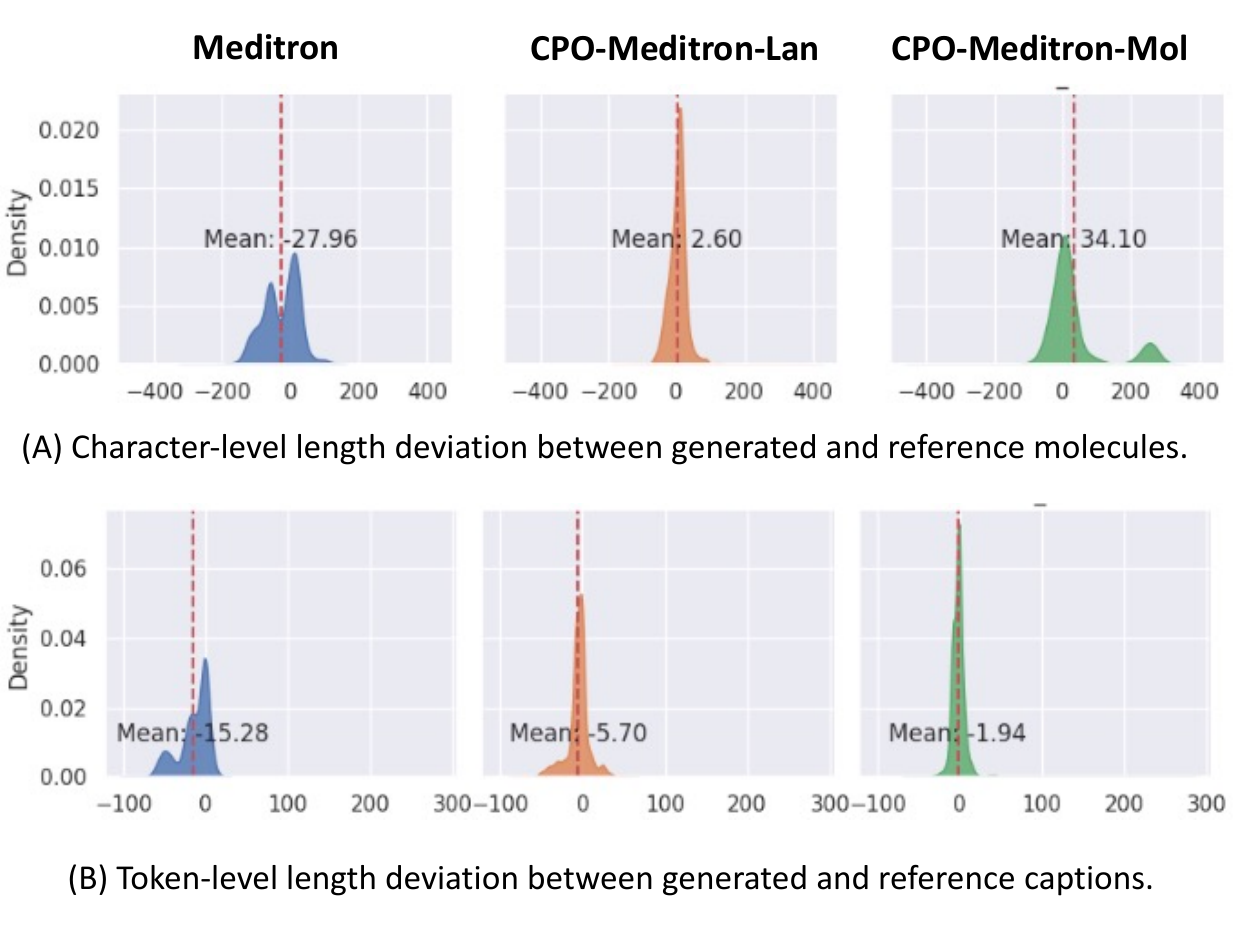}
\caption{Length-bias across different LLMs.} 
\label{fig:bias}
\end{figure}

For the language-to-molecule task, we observed that Meditron and Meditron$_{\overrightarrow{Mol.}}$ generated significantly shorter and longer outputs, respectively (see Fig.~\ref{fig:bias}). In contrast, Meditron$_{\overrightarrow{Lan.}}$ did not exhibit any length bias, producing outputs similar in length to the actual ones. Conversely, for the molecule-to-language task, our models did not show any significant length bias, while Meditron, trained on the entire dataset, generated significantly shorter answers against references.

\section{Conclusion}
This work address training efficacy and the out-of-distribution
problem for automatic language-molecule translation. We train models using only 10\% of available data and deploying contrastive preference optimisation which avoids generating translations that are merely adequate but not perfect. We achieve significant improvement in performance when compared with models trained on extensive in and out-of-the-distribution data. Finally, we propose a fine-grained, domain-agnostic evaluation method to assess hallucination in LLMs. Our models show superior factual consistency for caption generation and character-level n-gram overlaps for molecule generation.

\bibliography{custom}

\appendix

\section{Language-molecule Translation Instructions}
\label{appx:instructions}

\begin{figure}[h!]
\centering
\begin{tcolorbox}[colback=gray!4!white,colframe=black!75!black,fontupper=\small]
  Below is an instruction that describes a task, paired with an input that provides further context.\\
  Write a response that appropriately completes the request. \\ \\
  \#\#\# Instruction: You are a researcher. You can come up captions based on your existing knowledge. \\ Captions are given against the following input. You should be as detailed as possible. \\ \\
  \#\#\# Input: Molecule: \{\textcolor{blue}{source molecule}\} \\
  In that molecule, could you formulate a caption about? \\ \\ \\ 
  \#\#\# Response:\{\textcolor{blue}{target caption}\}
\end{tcolorbox}

\caption{Instruction for molecule to language translation, i.e., $M \rightarrow L$}
\end{figure}

\begin{figure}[h!]
\centering
\begin{tcolorbox}[colback=gray!4!white,colframe=black!75!black,fontupper=\small]
  Below is an instruction that describes a task, paired with an input that provides further context.\\
  Write a response that appropriately completes the request. \\ \\
  \#\#\# Instruction: You are a researcher. You can come up molecule smile strings based on your existing knowledge. \\
  Molecule smile strings are given against the following input. You should be as detailed as possible.\\ \\
  \#\#\# Input: Caption: \{\textcolor{blue}{source caption}\} \\
  In that caption, could you generate a molecule smile string? \\ \\ \\ 
  \#\#\# Response: \{\textcolor{blue}{target molecule}\}
\end{tcolorbox}

\caption{Instruction for language to molecule translation, i.e., $L \rightarrow M$}
\end{figure}

\section{Training Effectiveness and Efficiency}
\label{appx:effect_effic}

\begin{figure}[hbt!]
\centering
\includegraphics[width=.5\textwidth]{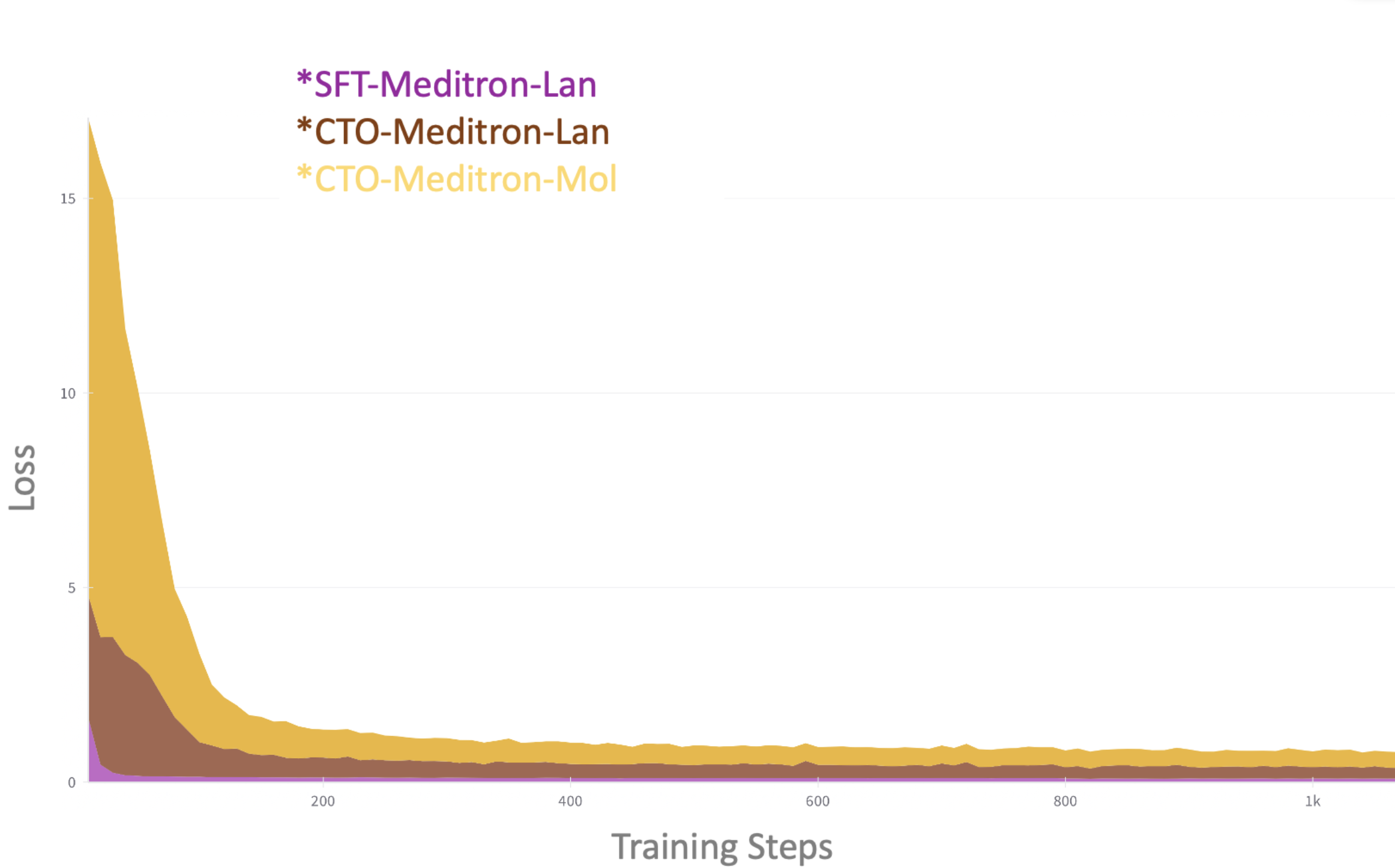}
\caption{Training convergence} 
\end{figure}

\begin{figure}[hbt!]
\centering
\includegraphics[width=.5\textwidth]{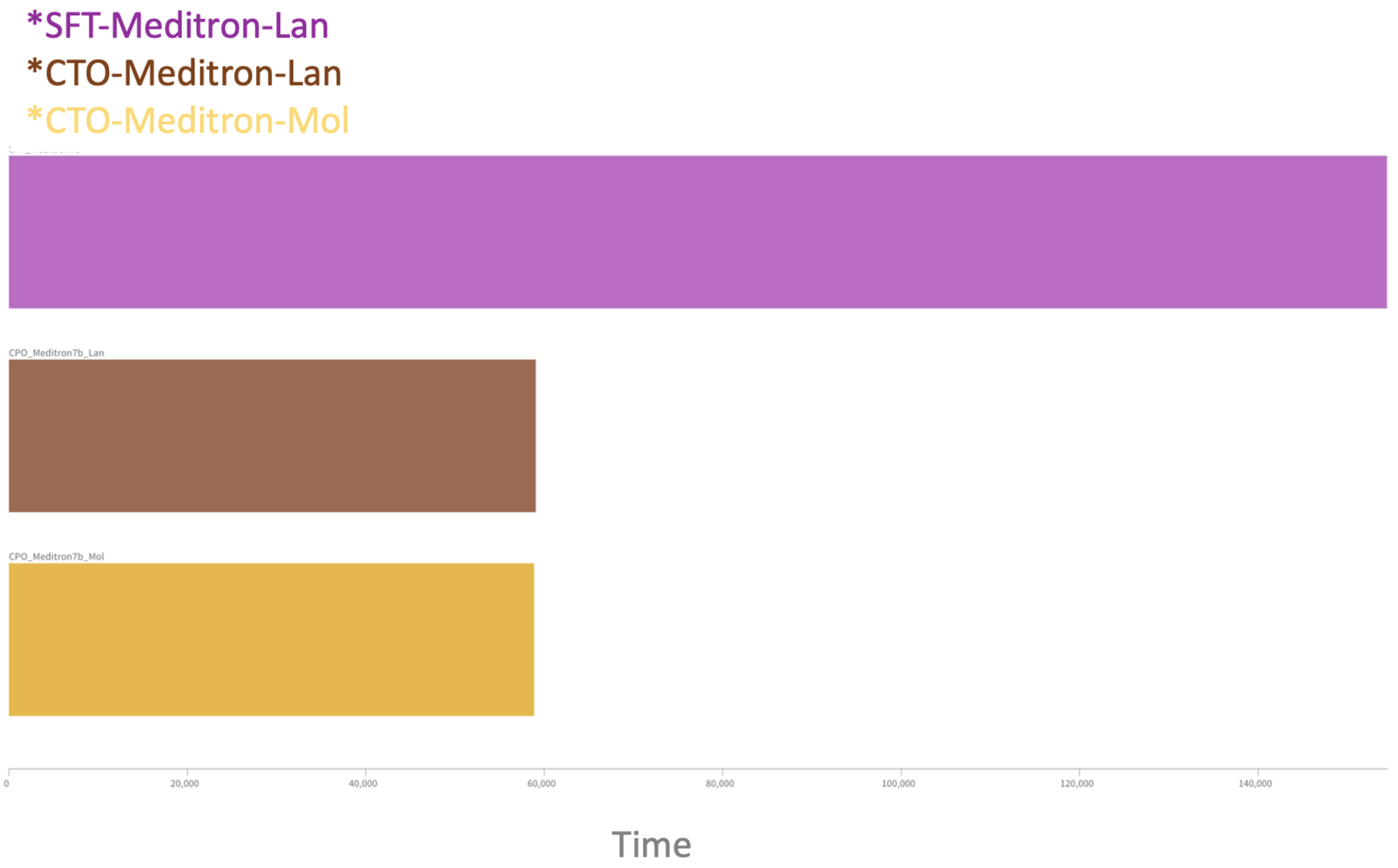}
\caption{Training efficiency} 

\end{figure}

\begin{figure}[hbt!]
\centering
\includegraphics[width=.5\textwidth]{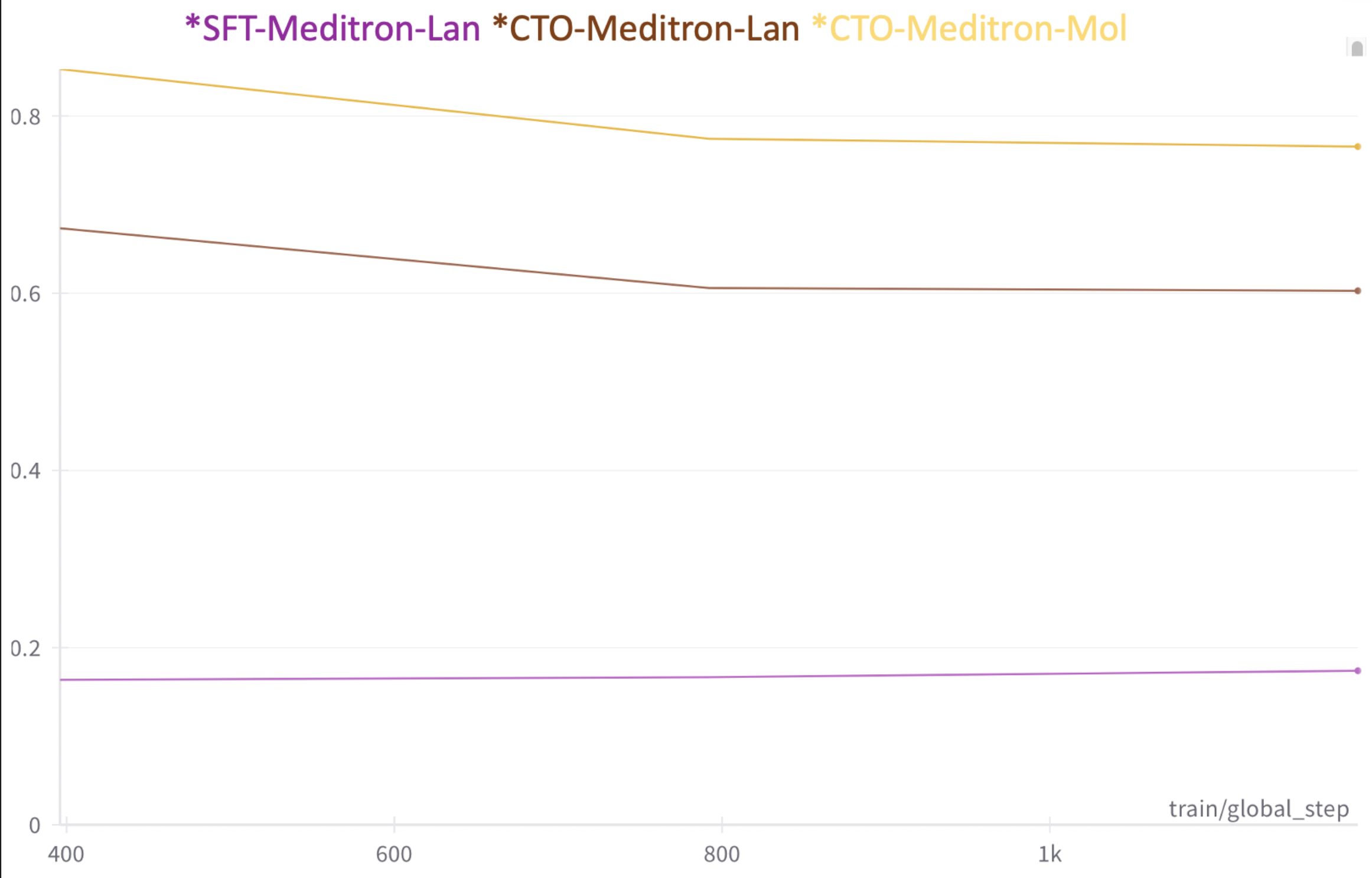}
\caption{Validation loss} 
\end{figure}







\end{document}